\documentclass{llncs}

\usepackage{graphicx}
\usepackage[T1]{fontenc}
\usepackage[latin1]{inputenc}

\begin{document}

\title{Combining Supervised and Unsupervised Learning for GIS Classification}

\author{Juan-Manuel Torres-Moreno\inst{1} \and Laurent Bougrain\inst{2} \and Fr{\'e}d{\'e}ric Alexandre\inst{2}}

\authorrunning{Torres-Moreno et al.}

\institute{Laboratoire Informatique d'Avignon\\
Université d'Avignon et des Pays de Vaucluse\\
BP 1228 84911 Avignon Cedex 09, France\\
 \and
 {\'E}quipe Cortex
 INRIA/LORIA Campus Scientifique \\BP 239 54506 Vand\oe uvre-lès-Nancy, Cedex, France\\
\email{juan-manuel.torres@univ-avignon.fr}
}

\maketitle              

\begin{abstract}
This paper presents a new hybrid learning algorithm for
unsupervised classification tasks. We combined Fuzzy c-means
learning algorithm and a supervised version of Minimerror to develop a
hybrid incremental strategy allowing unsupervised classifications.
We applied this new approach to a real-world database in order to
know if the information contained in unlabeled features of a
Geographic Information System (GIS), allows to well classify it.
Finally, we compared our results to a classical supervised classification
obtained by a multilayer perceptron.
\end{abstract}
{\bf Keywords :} Minimerror, Hybrid methods, Classification, Unsupervised learning, Geographic Information System.

\section{Supervised and Unsupervised Learnings}
For a classification task, the learning is supervised if the
labels of the classes of the input patterns are given a priori by
a professor. A cost function calculates the difference between
desired and real outputs produced by a network, then, this
difference is minimized modifying the network's weights by a
learning rule. A supervised learning set $\mathcal{L}$ is
constitued by $P$ couples
$(\vec{\xi}^{\mu},\tau^{\mu}),\mu=1,...,P$, where $\vec{\xi}^\mu$
is the input pattern $\mu$ and $\tau^\mu=\pm 1$ its class.
$\vec{\xi}^{\mu}$ is a $N$-dimension vector, with numeric or
categoric values. If labels $\tau^{\mu}$ are not present in
$\mathcal{L}$, it may be used as unsupervised learning. Learning
is unsupervised when the object's class is not known in advance.
This learning is performed by extraction of intrinsic regularities
of patterns presented to the network. The number of neurons of the
output layer corresponds to the desired number of categories.
Therefore, the network develops its own representation of input
patterns, retaining the statistically redundant traits.
\section{Supervised Minimerror}
Minimerror algorithm \cite{gor:gre} performs correctly in binary
problems of high dimensionality \cite{tor:gor:1,tor:gor:2,tor3}.
The supervised version of Minimerror performs a binary
classification using the minimization of the cost function:
\begin{equation}
 \label{eq:cost}
 E = \frac{1}{2} \sum_{\mu=1}^P
 V \left( \frac{\tau^\mu \vec{w} \cdot \vec{\xi}^\mu}{2T \sqrt{N}}
 \right)
\end{equation}
with
\begin{equation}
 V(x) = 1-\tanh(x)
\end{equation}
Temperature $T$ defines an effective window width on both sides of
the separating hyperplane defined by $\vec{w}$. The derivative
$\frac{dV(x)}{dx}$ is vanishingly small outside this window.
Therefore, if the minimum cost (\ref{eq:cost}) is searched through
a gradient descent, only the patterns $\mu$ at a
\begin{equation}
 |\gamma^\mu| \equiv { |\vec{w} \cdot \vec{\xi}^\mu| \over \sqrt{N}} < 2T
\end{equation}
distance will contribute significantly to learning
\cite{gor:gre,tor:1}. Minimerror algorithm implements this
minimization starting at high temperature. The weights are
initialized with Hebb's rule, which is the minimum of
(\ref{eq:cost}) in the high temperature limit. Then, $T$ is slowly
decreased upon the successive iterations of the gradient descent
by a deterministic annealing, so that only the patterns within the
narrowing window of width 2T are effectively taken into account
for calculating the correction
\begin{equation}
  \delta \vec{w} = -\epsilon\ { \partial E \over \partial \vec{w}}
\end{equation}
at each time step, where $\epsilon$ is the learning rate. Thus,
the search of the hyperplane becomes more and more local as the
number of iterations increases. In practical implementations, it
was found that convergence is considerably speeded-up if patterns
already learned are considered at a lower temperature $T_L$ than
the not learned ones, $T_L < T$. Minimerror algorithm has three
free parameters: the learning rate $\epsilon$ of the gradient
descent, the temperature ratio $T_L/T$, and the annealing rate
$\delta T$ at which temperature is decreased. At convergence, a
last minimization with $T_L = T$ is performed. This algorithm has
been coupled with a incremental heuristics, NetLS [2,5], which
adds neurons in one hidden layer as learning function. Several
results \cite{tor:1,tor:gor:1,tor:gor:2} show that NetLS is very
powerful and gives small generalization errors comparable to other
methods.

\section{Unsupervised Minimerror}

A variation of Minimerror, Minimerror-S \cite{tor:1,tor:gor:1},
allows to obtain spherical separations on input's space. The
spherical separation used the same cost function (1), but a
spherical stability $\gamma_s$ is defined by:
\begin{equation}
  \gamma_s = || \vec{w} - \vec{\xi}|| - \rho^2
\end{equation}
where $\rho$ is a hyperspherical's radius centered on $\vec{w}$.
The pattern's class is $\tau=-1$ inside the sphere and $\tau=1$
elsewhere. Spherical separations make it possible to consider
unsupervised learning using the Minimerror's separating qualities.
Thus, a strategy of unsupervised growing was developed in Loria.
The algorithm starts by obtaining the distances between the
patterns. The Euclidean distance can be used to calculate them.
Once the established distances, we started to find the pair $\mu$
and $\nu$ of patterns with the smallest distance $\rho$. This
creates the first incremental kernel. We located the hypersphere's
center $\vec{w_0}$ at the middle of patterns $\mu$ et $\nu$:
\begin{equation}
 \vec{w_0} = \frac{(\vec{\xi}^{\mu} + \vec{\xi}^{\nu})}{2}
\end{equation}
The initial radius is fixed
\begin{equation}
 \rho_0 = \frac{3\rho}{2}
\end{equation}
to make enter a certain number of patterns in growing kernel.
Then, patterns are labeled $\tau=-1$ if they are inside or in the
border of the initial sphere, and $\tau=1$ if elsewhere.
Minimerror-S finds the hypersphere $\{\rho*,\vec{w*}\}$ that
better separates patterns. The internal representations are
$\sigma=-1$ if
$$-\frac{1}{\cosh^2(\gamma^{\mu})}<\frac{1}{2}$$
else $\sigma=1$. This makes
it possible to check if there are patterns with $\tau=1$ outside
but sufficiently close to the sphere ($\rho_1^*, \vec{w_1^*}$). In
this case, then it makes $\tau=-1$ for these patterns and it
learns them again, repeating the procedure for all patterns of
$\mathcal{L}$. At this time, it passes to another growing kernel
which will form a second class $\vec{w_2}$, calculating with
Minimerror-S ($\rho_2^*,\vec{w_2^*}$), and repeating the procedure
until there is no more patterns to classify. Finally it obtains K
classes. A pruning procedure can avoid having too many classes by
eliminating those with few elements (less than one number fixed in
advance). It is possible to introduce conditions at the border,
which are restrictions that prevent locating the hypersphere
center outside of the input's space. For certain problems this
strategy can be interesting. These restrictions are however
optional: if it makes too many learning errors, the algorithm
decides to neglect them and the center and radius of separating
spheres can diverge.

\section{The Unsupervised Algorithm Fuzzy
c-means}

This algorithm \cite{bez,deg} allows us to obtain a clusterisation
of patterns with a fuzzy approach. Fuzzy c-means minimizes the sum
of the squared errors with the following conditions:
\begin{eqnarray}
 \sum_{k=1}^c m_{ik}& = & 1 ; \sum_{i=1}^n m_{ik} > 0 ; m_{ik} \in  {0,1}\\  & & i=1,2,\ldots,n ; k=1,2,\ldots,c
\end{eqnarray}
The objective function is defined by
\begin{equation}
 \label{eq:fuzzy2} J = \sum_{i=1}^n \sum_{k=1}^c  m^{\phi}_{ik}d^{2}(\xi_i,c_k)
\end{equation}
where $n$ is the number of patterns, $c$ is the desired number of
classes, $c_k$ is the centroid vector of class K, $\vec{\xi_i}$ is
a pattern $i$ and $d^2(\xi_i, c_k)$ is the square of the distance
between patterns $\xi_i$ and $c_k$, in agreement with a definition
of unspecified distance, which to simplify, we will indicate by
$d^2(\xi_i,c_k)$. $\phi$ is a fuzzy parameter, a value in $[2,
\infty)$, which determines the fuzzyfication of the final
solution, i.e., it controls the overlapping between the classes.
If $\phi=1$, the solution is a hard partition. If $\phi \to
\infty$ the solution approaches the maximum of fuzzyfication and
all the classes are likely to merge in only one. The minimization
of the objective function  $J$ provides the solution for the
membership function (6):

\begin{equation}  \label{eq:fuzzy3}  m_{ik}  =
\frac{d_{ik}^{2/{\phi-1}}}{\sum_{j=1}^c d_{ij}^{2/{\phi-1}}} ;
i=1,\ldots,n ;  k=1,\ldots,c;
\end{equation}

where:

\begin{equation}
 \label{eq:fussy3}
 c_{k}   =  \frac{\sum_{i=1}^n
m_{ik}^{\phi}x_{i}}{\sum_{i=1}^n m_{ik}^{\phi}} ; k=1,\ldots,c
\end{equation}

The fuzzy c-means algorithm is:
\begin{enumerate}
 \item Let the class number $k$, with $1<k<n$.
 \item Let a value of fuzzy parameter $f>2$.
 \item To choix a suitable distance definition in input's space. That may be euclidean distance
 and then $d^2(x_i,c_k) = || x_i - c_k ||^2$.
 \item To choix a value for stop criterium $\epsilon$ ($\epsilon = 0.001$ is a
 suitable convergence).
\item Let $M = M_{0}$, for pattern with random values or with values from a hard partition of k-means.
\item In iteration $t=1,2,3,...$ (re) calculate $C=C_t$ using \ref{eq:fussy3} and $M_{t-1}$.
\item Re-calculate $M=M_t$ using  equation \ref{eq:fuzzy2} and $C_t$.
\item To compare $M_t$ and $M_{t-1}$ with a suitable matrix norme. If $||M_{t} - M_{t-1}|| < \epsilon$ then stop else go to 6.
\end{enumerate}

\section{A Hybrid Strategy}
In spite of the supervised Minimerror's simplicity, the number of
classes  obtained is sometimes too high. Thus, we chose a combined
strategy: a first unsupervised hidden layer calculates the
centroids with Fuzzy c-means algorithm. As input we have $P$
unlabeled patterns of learning set $\mathcal{L}$. Then Supervised
Minimerror finds spherical separations well adapted to maximize
the stability of the patterns. The input is the same $\mathcal{L}$
set, but labeled by Fuzzy c-means. In this way, the number of
classes can be selected in advance.

\section{Deposit Prospection Experiment}

The mineral resources division of the French geological survey
(BRGM \cite{htt:2}) develops continent-scale Geographic
Information System (GIS), which support metallogenic research.
This difficult real-world problem constitutes a tool for decision
making. The understanding of the formation of metals such as gold,
copper or silver is not good enough and a lot of patterns
describing a site are available including the size of the deposit
for various metals. In this study, we will focus on a GIS which
covers all the Andes and two classes : deposit and barren. A
deposit is an economically exploitable mineral concentration
\cite{mic}. The concentration factor corresponds to the rate of
enrichment in a chemical element, i.e. to the relationship between
its average content of exploitation and its abundance in the
earth's crust. Geologists oppose to the concept of deposit the one
of barren. Actually, for the interpretation of the results of
generalization, it is necessary to enter the number of sites well
classified in each category to be able to answer the question: Is
this a deposit or a barren ? In our study, a deposit will be
defined as a site (represented by a pattern) that contains at
least one metal and a barren by a site without any metal. Then,
the classes \textit{deposit} and \textit{barren} will be used from
now on. The database we used contains 641 patterns, 398 examples
of deposits and 343 examples of barrens.

\subsection{Study of the Attributes}

\begin{figure}[h]
\centering
  \includegraphics[width=20pc]{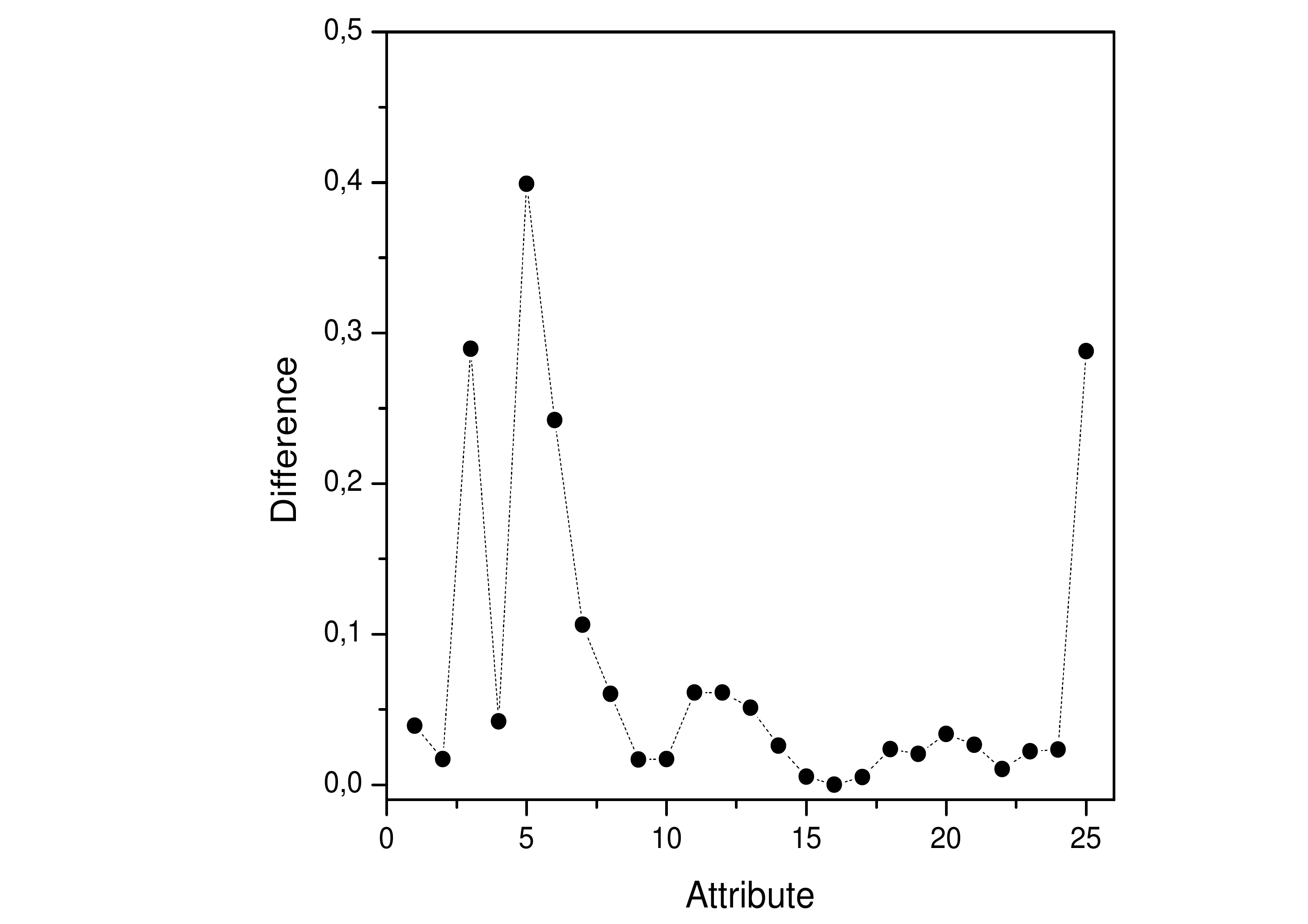}   
\caption{Mean squared differences of the average patterns.   }
\label{fig:dif}
\end{figure}

The original databases have 25 attributes, 8 qualitative and 17
quantitative, such as the position of a deposit, the type and age
of the country rock hosting the deposit, the proximity of the
deposit to a fault zone distinguished by its orientation in map
view, density and focal depth of earthquakes immediately below the
deposit, proximity of active volcanoes, geometry of the subduction
zone etc. We made a statistical study to determine the importance
of each variable. We calculated for each attribute the average of
\textit{deposit} and \textit{barren} patterns, in order to
determine which attributes were relevant for discriminating the
patterns (figure \ref{fig:dif}). There are some attributes (15,
16, 17 or 22, among others) that are not relevant. On the other
hand, the attributes 3, 5, 6 and 25 are rather discriminating. It
is interesting to know how the choice of attributes influences the
learning and specially the generalization tasks. Therefore, we
created 11 databases with different combinations of attributes.
Table 1 shows the number of qualitative and quantitative
attributes, and the dimension for each  database used.

\begin{table}[h]
\begin{center}
\begin{tabular}{|c|l c c c|}
\hline Database &    Attributes Used        & Qual.      & Quant.    & N \\
\hline\hline
I           &    1 to 25                & 8 & 17 & 25 \\
II          &    1 to 8                 & 8 & 0  & 8  \\
III & 9 to 25                           & 0 & 17 & 17 \\
IV  & 11,12,13,14                       & 0 & 4  & 4  \\
V   & 11,12,13,25                       & 0 & 4  & 4  \\ VI  & 3,5,6,7                           & 4 & 0  & 4  \\ \textbf{VII} & \textbf{11,12,13,14,25}  & \textbf{0} & \textbf{5}  & \textbf{5}  \\ VIII& 11,12,13,20,25                    & 0 & 5  & 5  \\ IX  & 3,5,6,7,11,12,13,25               & 4 & 4  & 8  \\ X   & 11,12,13,14,18,19,20,21,23,24     & 0 & 10 & 10 \\ XI  & 11,12,13,14,18,19,20,21,23,24,25  & 0 & 11 & 11 \\ [2pt]
\hline
\end{tabular}
\end{center}
\caption{Andes GIS learning databases used. }
\end{table}

\subsection{Data Preprocessing and  \textit{deposit/barren} Approach}
The range of the attributes is extremely broad. In order to
homogenize them,  a standardization  of quantitative attributes is
suitable. A data preprocessing is needed for the correct
functioning of the neural network. Thus, for each continuous
variable, the standardization calculates the average and standard
deviation. Then, the variable was centered and the values divided
by the standard deviation. The qualitative attributes are not
modified. The standardized corpus was divided in learning and test
sets. The sets consist of  randomly selected patterns from the
whole corpus. Learning sets of 10\%  (64 patterns) to 95\%  (577
patterns) of the original database (641 patterns) were generated.
The complement was selected as test set.
There are $N$ input neurons in the network, depending on the
database dimension. The unsupervised part of the network, Fuzzy
c-means, must find two classes: \textit{deposit} and
\textit{barren}. Minimerror will find the best hyperspherical
separator for each class. In the same condition, a multilayer
perceptron with 10 neurons on a single hidden layer obtains up to
77\% of correct classification.

\section{Results}
\begin{figure}[h]
\centering
  \includegraphics[width=20pc]{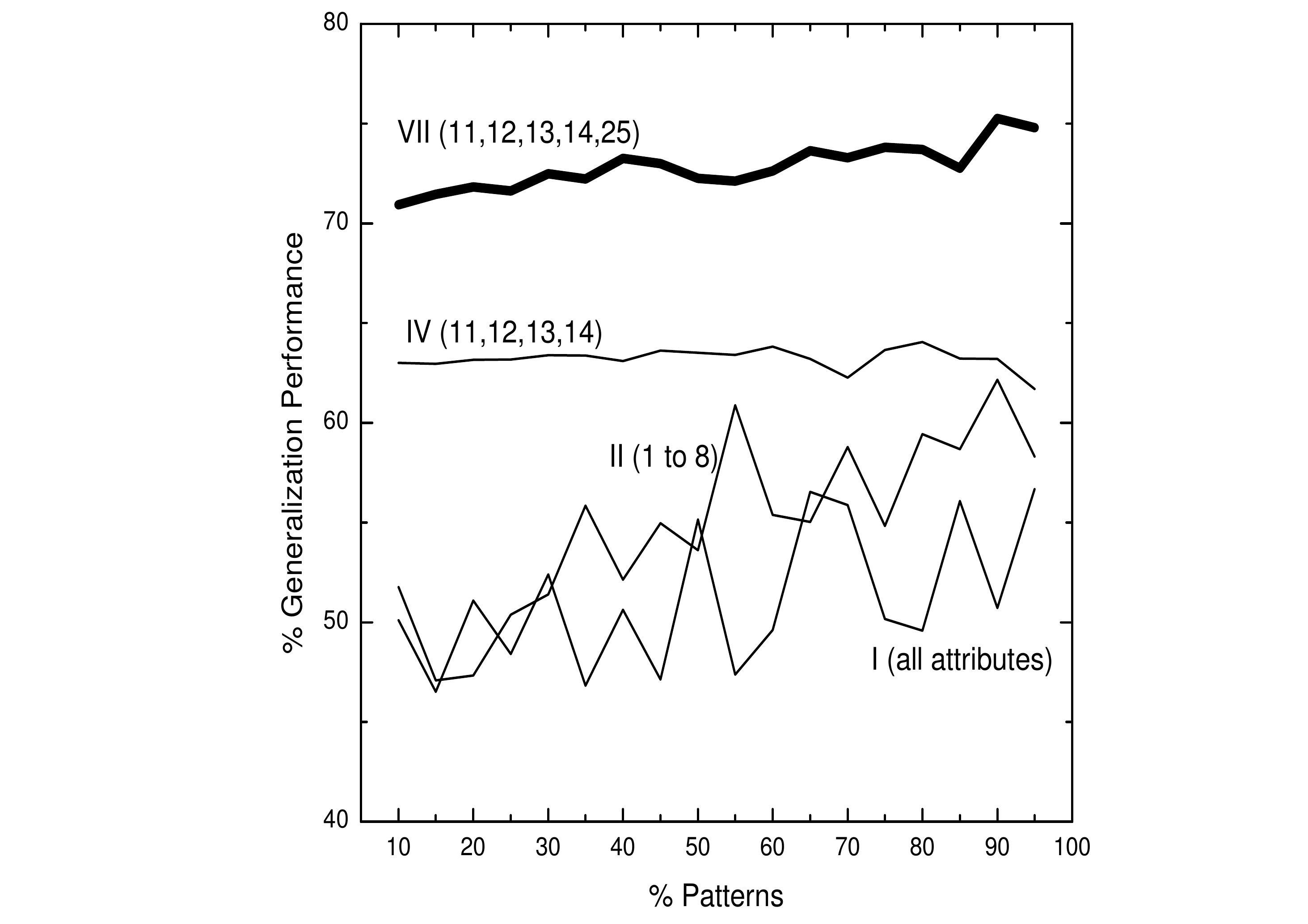}
\caption{Generalization performances  according to the learning set size obtained by the hybrid model with various databases.}
\label{fig:learngeneral}
\end{figure}

\begin{figure}[h]
\centering
\includegraphics[width=20pc]{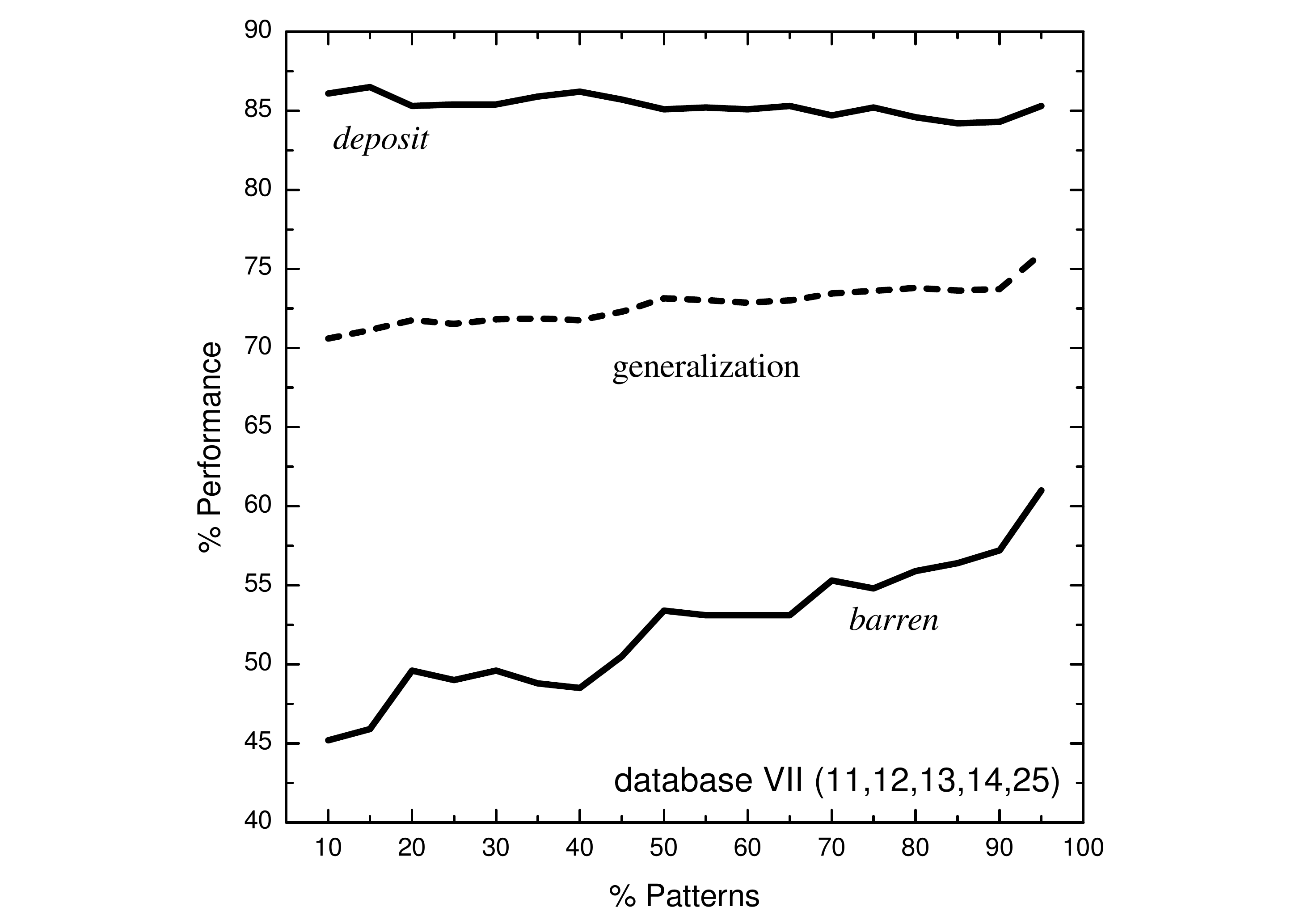}
\caption{\textit{deposit}/\textit{barren} discrimination
performances in generalization according to the learning set size
(100 tests) obtained by the hybrid model with the database VII.}
\label{fig:100test}
\end{figure}

 Classification performance corresponded to the percentage of well
 classified situations. Learning and generalization discrimination of \textit{deposit} and \textit{barren} were obtained for all learning databases. Database \textbf{VII} (including only few quantitative attributes) had the best learning and generalization performances in comparison to the other databases. When using all the attributes, the performances fell. Figure \ref{fig:learngeneral} shows some results of this behavior. Based on this information, we kept this database to perform 100 random tests. The capacity of discrimination between \textit{deposit} and \textit{barren}, according to the percentage of learned patterns is shown in figure \ref{fig:100test}. The \textit{deposit} class detection is quite higher than the \textit{barren} class.  We note  that the detection of gold, argent and copper remain quite  precise, bet, that of the molybdenum is rather poor. This can be  explained according to the weak presence of this metal.

\section{Conclusion}

We developed a variation of Minimerror for unsupervised
classification with hyperspherical separations. The hybrid
combination of Minimerror and Fuzzy c-means proved to be the most
promising. This strategy applied to real-world database, allowed
us to predict in a rather satisfactory way if a site could be
identified or not as a deposit. The 75\%  value obtained for the
well classified patterns with this unsupervised/supervised
algorithm is comparable to the values obtained with other
classical supervised methods. This also shows the discriminating
capacity of the descriptive attributes that we selected as the
most suitable for this two-class problem. Finally, according to
the figure 3, we should be able to obtain a significant
improvement of the performance just increasing the number of
examples. Additional studies must be made to determine more
accurately other relevant attributes, as well as to perform hybrid
learning multi-class tasks.

\section*{Acknowledgement}
  This research was supported by the \textsl{Bureau des Recherches G{\'e}ologique et Mini{\`e}re (BRGM)}, France.

\end{document}